\documentclass{article} % For LaTeX2e
\usepackage{iclr2021_conference,times}

\usepackage{amsmath,amsfonts,bm}

\def\eqref#1{equation~\ref{#1}}
\def\1{\bm{1}}

\DeclareMathAlphabet{\mathsfit}{\encodingdefault}{\sfdefault}{m}{sl}
\SetMathAlphabet{\mathsfit}{bold}{\encodingdefault}{\sfdefault}{bx}{n}

\usepackage{hyperref}
\usepackage{url}
\usepackage{graphicx}
\usepackage{caption}
\usepackage{subcaption}
\usepackage{float}

\title{Exploring the Similarity of Representations in Model-Agnostic Meta-Learning}

\author{Thomas Goerttler\textsuperscript{1} \& Klaus Obermayer\textsuperscript{1,2} \\
\textsuperscript{1}Technische Universität Berlin, Chair of Neural Information Processing, Germany\\
\textsuperscript{2}Bernstein Center for Computational Neuroscience Berlin, Germany\\
Correspondence: \texttt{thomas.goerttler@ni.tu-berlin.de} \\
}

\iclrfinalcopy % Uncomment for camera-ready version, but NOT for submission.
\begin{document}

\maketitle

\begin{abstract}
In past years model-agnostic meta-learning (MAML) has been one of the most promising approaches in meta-learning. It can be applied to different kinds of problems, e.g., reinforcement learning, but also shows good results on few-shot learning tasks. 
Besides their tremendous success in these tasks, it has still not been fully revealed yet, why it works so well. Recent work proposes that MAML rather reuses features than rapidly learns. In this paper, we want to inspire a deeper understanding of this question by analyzing MAML's representation.
We apply representation similarity analysis (RSA), a well-established method in neuroscience, to the few-shot learning instantiation of MAML. 
Although some part of our analysis supports their general results that feature reuse is predominant, we also reveal arguments against their conclusion. The similarity-increase of layers closer to the input layers arises from the learning task itself and not from the model. In addition, the representations after inner gradient steps make a broader change to the representation than the changes during meta-training.
\end{abstract}

\section{Introduction}\label{sec:introduction}

In recent years, meta-learning, also learning to learn \citep{DBLP:books/sp/98/ThrunP98}, has aroused interest in different fields of machine learning research. Meta-learning tries to use past experiences of similar problems to acquire a prior over model parameters or the learning procedure. Several researchers have used meta-learning to tackle the problem of few-shot learning \citep{DBLP:conf/icml/SantoroBBWL16, DBLP:conf/iclr/RaviL17, DBLP:conf/icml/FinnAL17} where novel tasks are trained only by seeing a few examples of each possible class. The models have to adapt quickly to the new problem task. Traditional supervised models fail because they cannot generalize on such a few examples and either do not learn the task or quickly overfit the training data.

Next to recurrent recurrent \citep{DBLP:conf/icann/HochreiterYC01} and attention-based  \citep{DBLP:conf/nips/VinyalsBLKW16} models, there exist also alternatives with a bi-level meta-learning setup e.g. MAML \citep{DBLP:conf/icml/FinnAL17}.
Despite its success, it is still not fully revealed why and also how MAML learns so well.
In \citet{DBLP:conf/iclr/RaghuRBV20} the question is proposed whether MAML successfully solves few-shot learning tasks mainly because of feature reuse or rapid learning. Feature reuse means that the meta-initialization is already so good that it does not have to be changed anymore significantly for the specific tasks. In contrast, rapid learning means that efficient significant changes are still done. By freezing a lot of layers and comparing the different layers by applying canonical correlation analysis (CCA) \citep{DBLP:conf/nips/MorcosRB18}, and centered kernel alignment (CKA) \citep{DBLP:conf/icml/Kornblith0LH19}, they come to the answer that MAML rather reuses features.

We analyze the representation of few-shot learning models during training with representation similarity analysis (RSA) \citep{kriegeskorte2008representational} and reveal further insights into their representations.
Similar to \citet{DBLP:conf/iclr/RaghuRBV20}, we observe that the similarity of trained activations to activations before training is very high in early layers. However, we think that this is caused rather by the task itself than by the MAML learning procedure, as the same effect can be observed in CNNs for standard supervised learning. 
Moreover, our analysis also reveals that inner gradient steps lead to a more significant change in the representation than the updates in meta-training.
Besides, we observe that most of the inner step optimization happens in the first gradient step.

%Some meta-learning algorithms use recurrent \citep{DBLP:conf/icann/HochreiterYC01, DBLP:conf/icml/SantoroBBWL16, DBLP:conf/iclr/RaviL17} or attention-based \citep{DBLP:conf/nips/VinyalsBLKW16, DBLP:conf/iclr/MishraR0A18} models trained via a meta-learning objective. %These techniques attempt to encapsulate the learned learning procedure in the parameters of the neural net.
%An alternative approach has been proposed in by \citet{DBLP:conf/icml/MaclaurinDA15} and \citet{DBLP:conf/icml/FinnAL17} where a bi-level meta-learning setup exists. The \textit{inner} optimization represents an adaptation to a given task, whereas the \textit{outer} objective is the meta-training objective. Especially model-agnostic meta-learning (MAML) has been successful, and numerous extensions on them have been proposed to speed it up and improve its effectiveness \citep{DBLP:conf/icml/FinnAL17, DBLP:journals/corr/abs-1803-02999, DBLP:journals/corr/abs-1909-04630}.

\section{Related Work}\label{sec:related_work}

%By using this formulation, we can learn the initial parameters of a model, which leads to fast adaptation and generalization when optimizing from this initialization.

Model-agnostic meta-learning (MAML) addresses the
general problem of meta-learning \citep{DBLP:books/sp/98/ThrunP98}, which includes few-shot learning. Next to recurrent-based \citep{DBLP:conf/icann/HochreiterYC01, DBLP:conf/icml/SantoroBBWL16, DBLP:conf/iclr/RaviL17}, and attention-based \citep{DBLP:conf/nips/VinyalsBLKW16, DBLP:conf/iclr/MishraR0A18} meta-learning models, there also exist model-agnostic approaches \citep{DBLP:conf/icml/MaclaurinDA15, DBLP:conf/icml/FinnAL17}.
Model-agnostic meta-learning \citep{DBLP:conf/icml/FinnAL17} is a model-agnostic approach, which means it does not require a specific model. Since MAML requires second-order derivatives, there are several approaches that approximate the second-order derivatives, e.g., FOMAML \citep{DBLP:conf/icml/FinnAL17}, Reptile \citep{DBLP:journals/corr/abs-1803-02999}, and iMAML \citep{DBLP:journals/corr/abs-1909-04630} to overcome the need of calculating them.

Representation similarity analysis (RSA) is widely used in computational neuroscience to compare a computational or behavioral model with the brain response \citep{kriegeskorte2008representational}. In \citet{10.1371/journal.pcbi.1003915}, several (un)supervised vision models are used to show that supervised models are better for explaining IT cortical than unsupervised ones. In \citet{DBLP:journals/neuroimage/CichyKPO17} the correlation of dynamics of the visual system is correlated with deep networks.
RSA is also used in pure computational models \citep{DBLP:journals/ficn/McClureK16, DBLP:conf/cvpr/DwivediR19, mehrer}:
\citet{DBLP:journals/ficn/McClureK16} use RSA as a loss function for knowledge distillation, \citet{DBLP:conf/cvpr/DwivediR19} use it as a measure for the similarity between vision tasks, and \citet{mehrer} analyze the consistency of neural networks with RSA.
\citet{DBLP:conf/eccv/DwivediHCR20} propose a more general approach of comparing different activations, that RSA and CKA \citep{DBLP:conf/icml/Kornblith0LH19} - which is used in \citet{DBLP:conf/iclr/RaghuRBV20} to analyze MAML - are instantiations of.
\section{Analysis of Representation in Few-Shot Learning}

The idea of MAML \citep{DBLP:conf/icml/FinnAL17} is to learn an initialization of a model using exact second-order methods across tasks that are sampled from the same distribution. The optimization is done in a bi-level meta-learning setup. The meta-optimization is done in the outer loop and can be described as 
\begin{equation}\label{equ:outer}
\theta^* := \underset{\theta \in \Theta}{\mathrm{argmin}} \frac{1}{M} \sum_{i=1}^M \mathcal{L}(in(\theta, \mathcal{D}_i^{tr}), \mathcal{D}_i^{test})
\end{equation}
where $M$ is the number of task in a batch, $\mathcal{D}_i^{tr}$ and $\mathcal{D}_i^{test}$ are the training and test set of task $i$, $\mathcal{L}$ is the loss function and $in(\theta, \mathcal{D}_i^{tr})$ describes the inner loop. For every task $i$ in a batch, the neural network is initialized with $\theta$ and optimized for one or a few training steps of gradient descent to obtain $\phi_i$.
\begin{equation}
\phi_i \equiv in(\theta, \mathcal{D}_i^{tr}) = \theta - \alpha  \nabla_{\theta} \mathcal{L}(\theta, \mathcal{D}_i^{tr})
\end{equation}
is an example of only using one gradient step.

In this paper, we apply RSA to the representations which are obtained by training MAML on few-shot learning tasks to reveal dissimilarities.
These dissimilarities are obtained by comparing pair-wise dissimilarities (e.g., by Euclidean distances) between representations associated with each pair of inputs. This results in Representation Dissimilarity matrices (RDMs). In a second step, the Spearman’s correlation between different RDMs of different representations is calculated. Subtracting these from 1 leads to the dissimilarity scores of representations. 
In MAML, after every training step there exist a new initial representations parametrized by $\theta$ as well as new fine-tuned representations for every task $i$ parametrized by $\phi_i$ of the meta-test task which are obtained after inner loop updates on the training (support) $\mathcal{D}_i^{tr}$ set.
We analyze both the initial representation and the fine-tuned ones for several meta-test tasks.
Further details on the application of RSA are described in Appendix~\ref{app:rsa}. 

%Traditionally, the meta-test task is analyzed by the performance of the test (or query) set. As this query set differs across every test task, but RSA needs the same inputs aligned, we hold out one specific evaluation task to compare different models after different training steps.
%To guarantee that a test task does not overlap partially with the evaluation task, none of the different test tasks shares input samples with the evaluation task.
%In addition, all of the meta-test tasks we analyze do not have any class in common with the classes from other meta-test tasks.

In our experiments, we use the same architecture which was proposed in \citet{DBLP:conf/icml/FinnAL17}. In the analysis, we focus on 20-way 1-shot learning on the omniglot dataset, but results on other tasks (e.g., Mini-ImageNet) showed the same behaviour (see Appendix~\ref{app:miniimagenet}). Details on the model are described in Appendix~\ref{app:model}. How to access the code can be found in Appendix~\ref{app:code}.

Additionally, we also apply the same network architecutre on MNIST for standard supervised classication to observe the similarity there.

\section{Results}\label{sec:results}

\begin{figure}
     \centering
     \begin{subfigure}[b]{0.40\textwidth}
         \centering
         \includegraphics[width=\textwidth]{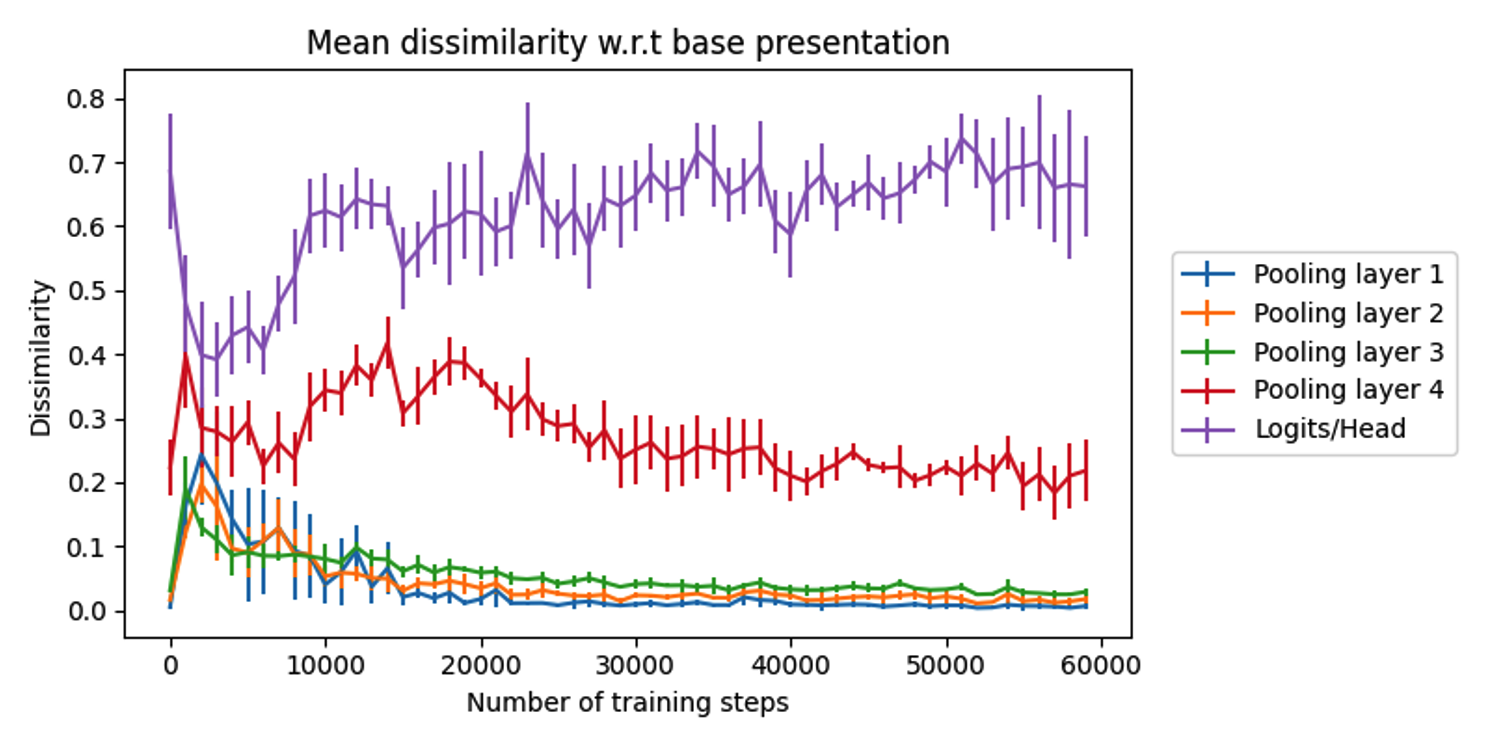}
         \caption{}
         \label{fig:dissimilarity_to_base}
     \end{subfigure}
     \hfill
     \begin{subfigure}[b]{0.40\textwidth}
         \centering
         \includegraphics[width=\textwidth]{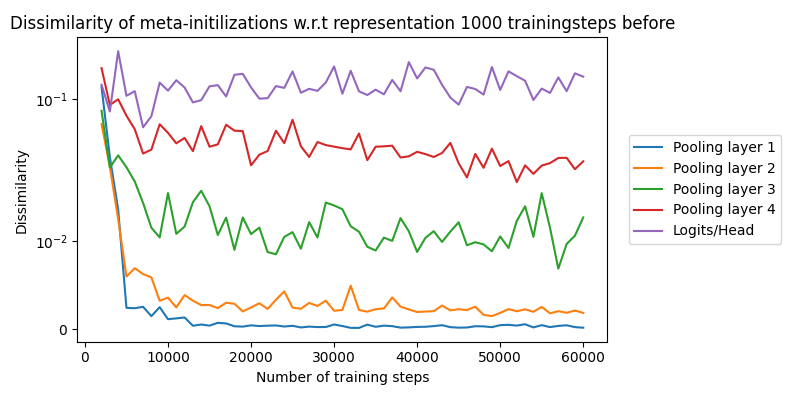}
         \caption{}
         \label{fig:dissimilarity}
     \end{subfigure}
     
    \caption{On (a), we see the mean dissimilarity (RSA with Euclidean distance) with error bars of the representations before finetuning compared to the random initilization at the beginning of meta-training of 50 test task. On (b), we see the dissimilarity of the representations before finetuning to the one 1000 training steps before (log-scaled). }
\end{figure}

\begin{figure}
     \centering
     \begin{subfigure}[b]{0.40\textwidth}
         \centering
         \includegraphics[width=\textwidth]{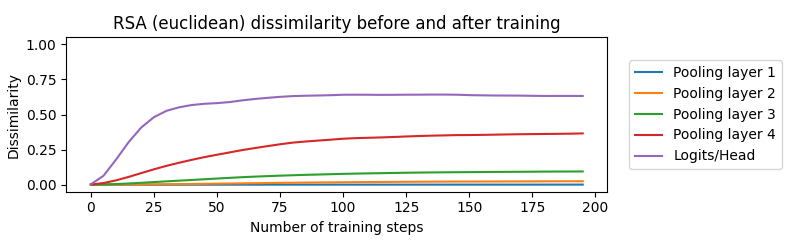}
         \caption{}
        \label{fig:class1}
     \end{subfigure}
     \hfill
     \begin{subfigure}[b]{0.40\textwidth}
         \centering
         \includegraphics[width=\textwidth]{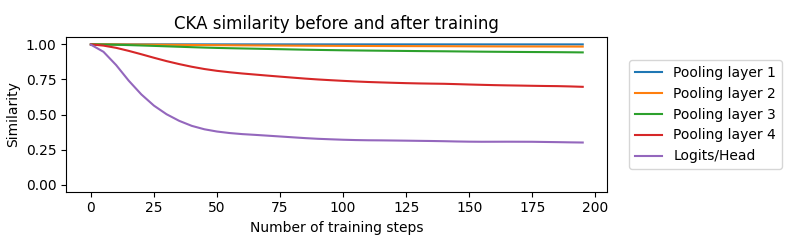}
         \caption{}
        \label{fig:class2}
     \end{subfigure}
       \caption{In this figure, the (dis)similarity of the MNIST experiments is seen. On (a), we use RSA and in (b) CKA as a (dis)similarity measure. Note that CKA measures the similarity, whereas RSA measures dissimilarity.}
        \label{fig:class}
\end{figure}

\paragraph{Comparision of similarity of early and late layers:}
In Figure~\ref{fig:dissimilarity_to_base}, we analyze the dissimilarity of the representations after 5 fine-tuning gradient steps on the test set of 50 meta-test tasks. After 60000 meta-training steps, we observe the same circumstance like \citet{DBLP:conf/iclr/RaghuRBV20}: The similarity of the representations of early layers to its initialization is really high, whereas especially in the last layer, the activations change a lot. This makes sense because the last layer has to adapt to the labels of the novel task. \citet{DBLP:conf/iclr/RaghuRBV20} conclude from this observation that almost no learning happens in the inner loop.
However, we also observe that this circumstance does not only hold after 60000 but also in earlier stages of the training process and even for the first ones.
In fact, this even holds for standard classification with now pre-training of the initial $\theta$. In a second experiment, we use the same network which is used for MAML on Omniglot for classification on MNIST (see Figure~\ref{fig:class}). 
Although the weights of early layers are completely random, the activations of early layers hardly change. To align with \citet{DBLP:conf/iclr/RaghuRBV20}, we also calculated the similarity with CKA, which shows the same but even with higher similarity in early layers.

From these experiments, we conclude that even if no feature reusing is possible as weights are random, during fine-tuning, the activation of early layers stays more similar to their initializations than in later layers. Instead of deriving from this that no learning happens, we think that this is probably more likely an artifact of the model itself (early layers are farther away from the output layer and get smaller errors and gradients), or the network architecture is too dense. Therefore, we think this observation is not property-specific to MAML.

\paragraph{Comparision of meta-training to fine-tuning:}

Figure \ref{fig:dissimilarity} shows the dissimilarity using RSA of the representation parametrized by $\theta$ of the five layers to the representation obtained by the network exactly 1000 training steps before.
After several thousand training steps, we see that the dissimilarity has converged to a value and then only fluctuates around it for the rest of the training time. %We can also observe that the earlier a layer in the network is, the more similar its representations to the one 1000 steps before are. 
The dissimilarity of representations from the first layers is close to $0$, whereas the last convolution layers' dissimilarity fluctuates around about $0.1$. This underlines the observation of before that there is more change in later layers than in early ones.

\begin{figure}[t]
     \centering
     \begin{subfigure}[b]{0.45\textwidth}
         \centering
         \includegraphics[width=\textwidth]{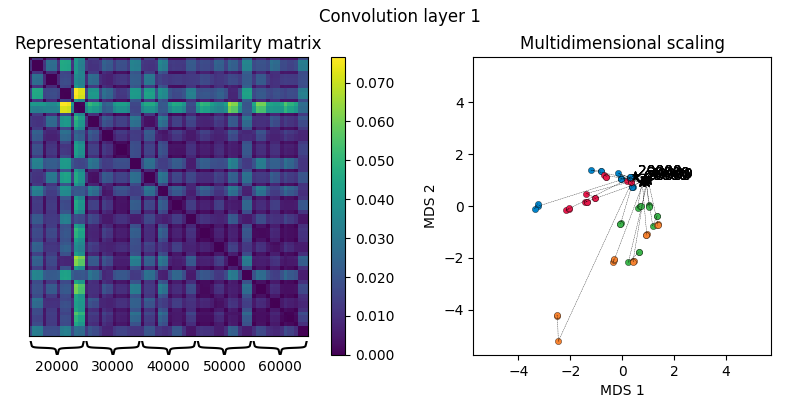}
         \caption{}
         \label{fig:subfig1}
     \end{subfigure}
     \hfill
     \begin{subfigure}[b]{0.45\textwidth}
         \centering
         \includegraphics[width=\textwidth]{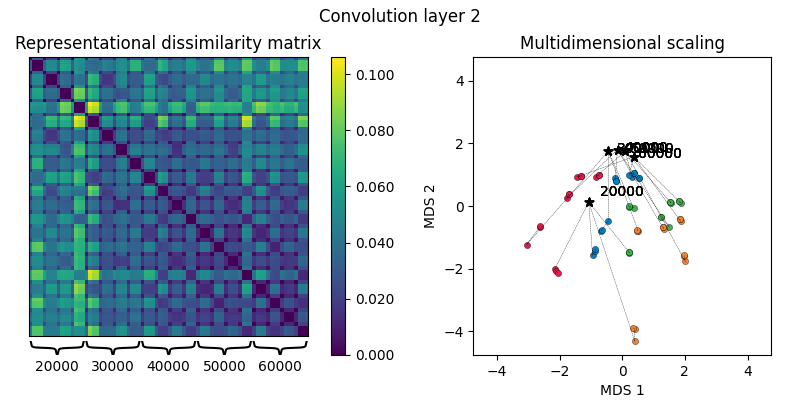}
         \caption{}
         \label{fig:subfig2}
     \end{subfigure}
     \hfill

     \begin{subfigure}[b]{0.45\textwidth}
         \centering
         \includegraphics[width=\textwidth]{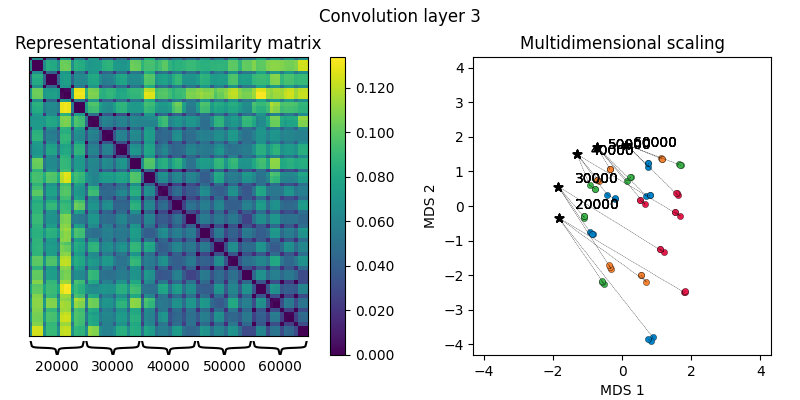}
         \caption{}
         \label{fig:subfig3}
     \end{subfigure}
     \hfill
     \begin{subfigure}[b]{0.45\textwidth}
         \centering
         \includegraphics[width=\textwidth]{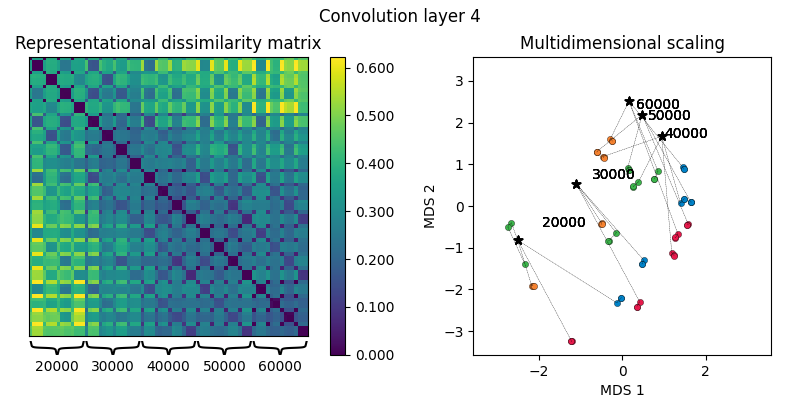}
         \caption{}
         \label{fig:subfig4}
     \end{subfigure}
     
        \caption{In this figure, the dissimilarity of the representations obtained by applying an evaluation task on the MAML model after 20000, 30000, 40000, 50000, and 60000 training steps are shown: on the left the RDM, on the right side after applying multidimensional scaling. For every step, we obtain the representation after updating for 0, 1, 5, and 10 inner steps connected by a dotted line.}% Each color is referring to the same task for every layer and every step. The results of the head layer can be found in Appendix~\ref{app:ominiglot}.}
         \label{fig:results}
\end{figure}

For further analysis, we sampled four tasks and evaluated the representation on a fifth one. The performance of them increasing in the first 10000 training iterations and after that only slowly increasing fluctuating around it (Further details in Appendix~\ref{app:ominiglot}).
%In Figure \ref{fig:performance} their performance after several training steps is shown. We observe they start fluctuating around a value after some thousand training steps increasing slightly. Also, the average performance across 600 tasks is rapidly increasing during the first 10000 iterations and then only slowly.
In Figure~\ref{fig:results} we see the results of applying RSA on the representation before and after (1, 5, and 10) inner loop gradient steps of the four selected tasks for all layers after 20000, 30000, 40000, 50000, and 60000 training steps. %As seen in Figure \ref{fig:performance}, the different models' performance is comparatively similar, although it is fluctuating quite a bit.
%First, we see that the dissimilarity scores in earlier convoltuion layers are smaller than in the later convolution layers. This underlines the theory that fine-tuning benefits rather from reusing features than from quickly learning.
We see that in every layer, all fine-tuned representations, even only after one inner step, are quite far away from their meta initialization by $\theta$. Especially for the early layers, the learned representations in meta-optimization are closer to the representations 10000 training steps before than to their corresponding fine-tuned query tasks. 
This points to a rapid adaptation to the novel situation also in early convolution layers.

Although the representations after fine-tuning are similar to their starting representation in general, compared to the changes of updating during meta optimization (see \ref{fig:dissimilarity}), the change is significantly larger. Interestingly the first inner step is responsible for most of the change in the representations. Trained for 5 inner steps, the representations after 5 or even 10 inner steps are not significant compared to the one after 1 step.% This can be seen by looking at the dotted line, which connects the representation after 1, 5, and 10 inner steps.
Therefore, we think that rapidly learning also happens in early layers, even if their representation stays more similar to its initialization than in later layers.

\section{Conclusion}

In this paper, we applied RSA on both the initial but also the fine-tuned representation of MAML. On the one hand, our analysis confirmed that the representations in early layers do not change a lot compared to in later layers.
However, this is also true in supervised classification. Therefore, this observation is not specific to MAML, but we think more likely caused by the architecture of neural networks.

In addition, we also showed that in early layers, the change of the representations in the inner loop of meta-testing is significantly larger than the representation change during training the meta-initialization. This reveals that MAML partially still quickly learns also in early layers. 

%Furthermore, most of the change in the representation seems to be done by only the first inner-step. 

In general, we think that the question of whether MAML reuses features or rapidly learns also depends on its definition. As there is no clear definition and measure for that, it is hard to agree finally.
Nevertheless, our analysis shows the potential of using RSA to analyze the representations of MAML. We think that more questions, e.g., if a test task aims to a similar optimum during training, can be discussed in the future by looking at the representations of MAML models.

\bibliographystyle{iclr2021_conference}
\bibliography{references}
\newpage

\appendix
\section{Model details}\label{app:model}

In our experiments, we use the same architecture proposed in \citet{DBLP:conf/icml/FinnAL17}. 
In our analysis, we focus on 20-way 1-shot learning on Omniglot in Section~\ref{sec:results} and in the Appendix \ref{app:miniimagenet} on 5-way 1-shot on Mini-ImageNet.

For both datasets, we have 4 convolution hidden layers with a 3x3 filter and 64 filters for Omniglot and 32 filters for Mini-ImageNet, batch normalization, and ReLU activations. For Omniglot, the dimension of the strides is 2x2; for Mini-ImageNet, we use 2x2 max pooling instead. The output layer is a dense layer fed in a softmax. The images of Omniglot are downsampled to 28x28.

The network are trained for 60000 training steps with a batch size of $16$ (Omniglot) and $4$ (Mini-ImageNet), $5$ inner loop updates, an inner learning rate of $0.1$ ($0.01$ for Mini-ImageNet), and a meta-learning rate of $0.001$ (AdamOptimizer).

For the MNIST experiement we use the same network as for Omniglot. Everything stays the same, only the batch size increases to 100 and we only traine for 200 training steps.

\section{Details of measuring the dissimilarity with RSA}\label{app:rsa}

\paragraph{RSA} Representation dissimilarity analysis is organized in two steps. In the first step, we compute a representation dissimilarity matrix (RDM) for every representation we want to compare. This is done by calculating the pair-wise dissimilarity associated with each pair of inputs resulting in a representations' gram matrix. The resulting matrix is a symmetric matrix with the dimension of the number of inputs we consider. The dissimilarities are originally calculated by $1-corr_{Pearson}(x,y)$ where $x$ and $y$ are representations; however, other (dis)similarity functions can be used. In our paper, we used the Euclidean distance.

In a second step, we again take pair-wise dissimilarities, but this time we take the flattened RDM matrices as the input obtained in the first step. For computational reasons, we only take the lower or upper triangle.
This time the dissimilarity is calculated by $1-corr_{Spearman}(x,y)$ where $x$ and $y$ are RDMs from step 1. 

\paragraph{Experimental details}
Traditionally, the meta-test task is analyzed by the performance of the test (or query) set. As this query set differs across every test task, but RSA needs the same inputs aligned, we hold out the same specific evaluation task to compare different models after different training steps.
In our experiment where we show the dissimilarity of four different tasks given the same initilization (Figure~~\ref{fig:results}), we use always the same hold out task to calculate the dissimilarity.

\section{Further result on Omniglot}\label{app:ominiglot}

\begin{figure}[h]
\centering
\includegraphics[width=0.45\textwidth]{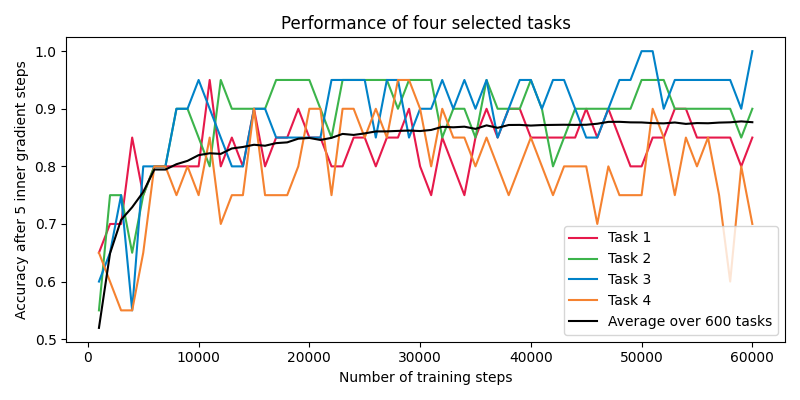}

\caption{On this figure, we see the performance of MAML of four sampled tasks and optimized for five iterations in the inner loop.}
\label{fig:performance}
\end{figure}

In Figure \ref{fig:performance} the performance of the four task of Figure~\ref{fig:results} after several training steps is shown. We observe they start fluctuating around a value after some thousand training steps increasing slightly. Also, the average performance across 600 tasks is rapidly increasing during the first 10000 iterations and then only slowly.

\begin{figure}[h]
\centering
\includegraphics[width=0.45\textwidth]{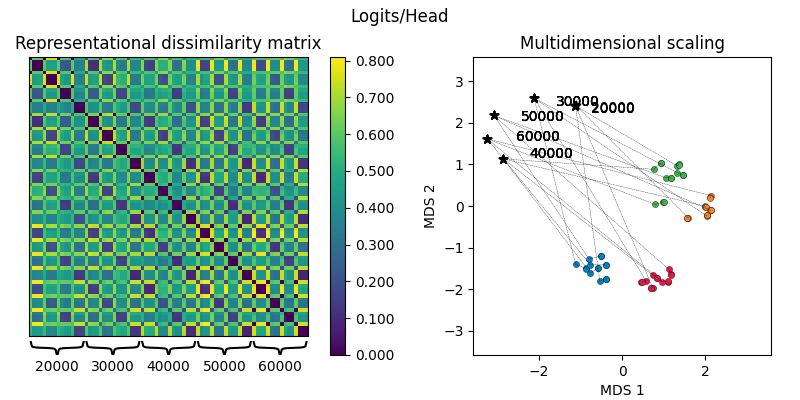}
\caption{In this figure we show the representations of the output layer of the experiement in Section~\ref{sec:results}}.
\label{fig:head}
\end{figure}

In Figure~\ref{fig:head} the representations of the head layer of the experiment on the omniglot dataset is shown. Not surprisingly we see, that every task tends to go the same region meaning that their representation is similar. This makes sense, as the output layer has to have very similar representations to classify similary.

\section{Results on Mini-Imagenet}\label{app:miniimagenet}

\begin{figure}[h]
     \centering
     \begin{subfigure}[b]{0.45\textwidth}
         \centering
         \includegraphics[width=\textwidth]{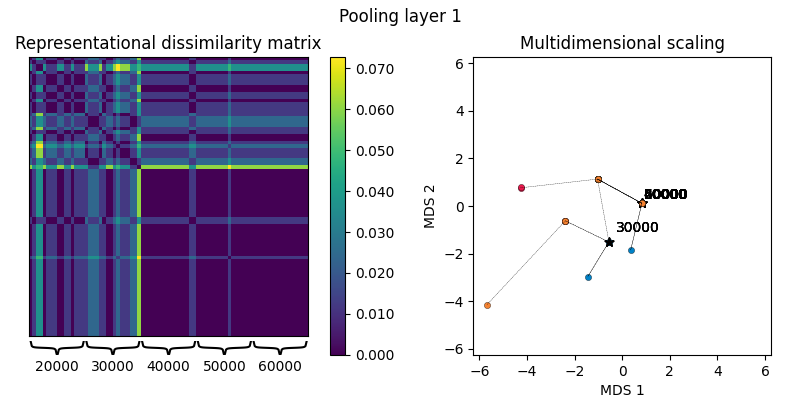}
         \caption{}
         \label{fig:subfig1}
     \end{subfigure}
     \hfill
     \begin{subfigure}[b]{0.45\textwidth}
         \centering
         \includegraphics[width=\textwidth]{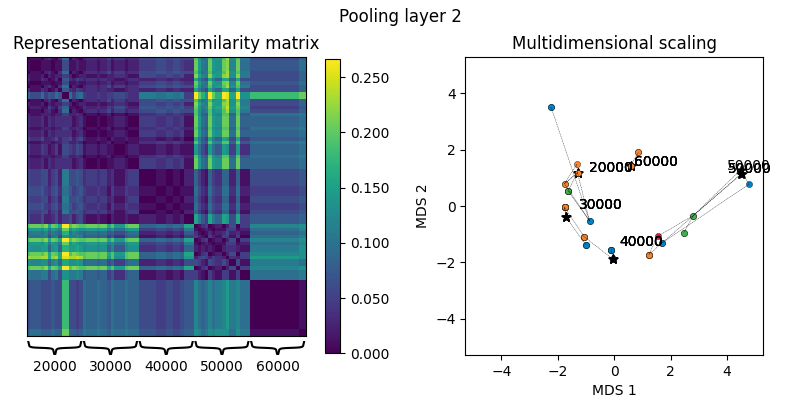}
         \caption{}
         \label{fig:subfig2}
     \end{subfigure}
     \hfill

     \begin{subfigure}[b]{0.45\textwidth}
         \centering
         \includegraphics[width=\textwidth]{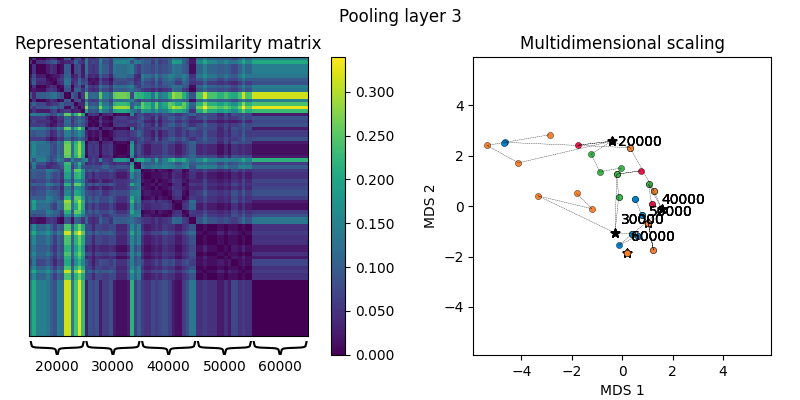}
         \caption{}
         \label{fig:subfig3}
     \end{subfigure}
     \hfill
     \begin{subfigure}[b]{0.45\textwidth}
         \centering
         \includegraphics[width=\textwidth]{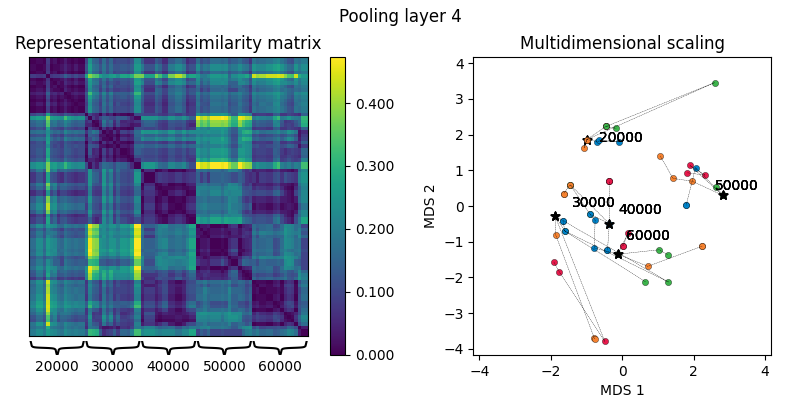}
         \caption{}
         \label{fig:subfig4}
     \end{subfigure}

     \begin{subfigure}[b]{0.45\textwidth}
         \centering
         \includegraphics[width=\textwidth]{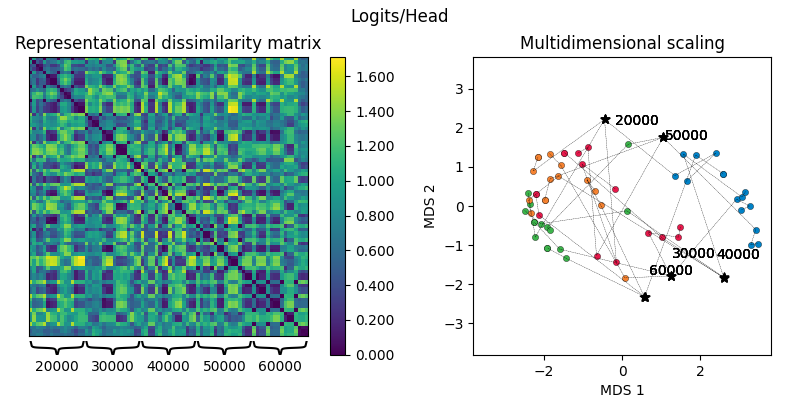}
         \caption{}
         \label{fig:subfig4}
     \end{subfigure}
     
        \caption{This figure shows the same as Figure~\ref{fig:results}, this time on Mini-ImageNet.}
         \label{fig:miniimagenet}
\end{figure}

We also repeated the experiment on a more difficult dataset Mini-ImageNet (see Figure~\ref{fig:miniimagenet}), and observed the same as for Omniglot.

%Another interesting observation is that, especially in later layers, the same tasks starting from different meta-initializations are heading to a similar representation region that is more similar to themselves than to initialization. 
%For the last layer, this makes sense as the loss function directly optimizes for it. 
%This might seem surprising because there could be many other possible representations as we still have to consider that we fine-tune only on 20 data samples per task. 
%This also leads to the result that inner optimization is valuable and leads to better representation than not training them.

\section{Access to the Code}\label{app:code}

The code to reproduce all the results is available on GitHub\footnote{https://github.com/ThomasGoerttler/similarity-analysis-of-maml}. For further details, please follow the instructions there. The MAML models are trained with the code from the original work on MAML \citep{DBLP:conf/icml/FinnAL17}\footnote{https://github.com/cbfinn/maml}.

\end{document}